%% file: main.tex
\documentclass[sigconf,review=false,anonymous=false]{acmart}

\renewcommand\footnotetextcopyrightpermission[1]{}

\AtBeginDocument{%
  \providecommand\BibTeX{{%
    \normalfont B\kern-0.5em{\scshape i\kern-0.25em b}\kern-0.8em\TeX}}}

\usepackage{paralist}
\usepackage{fullpage}
\usepackage{times}
\usepackage{fancyhdr,graphicx,amsmath}
\usepackage[ruled,vlined]{algorithm2e}
\usepackage{subcaption}
\include{pythonlisting}
\usepackage{bm}
\newcommand{\HBS}[1]{{\textcolor{black}{#1}}}

\begin{document}
\fancyhead{}

\title{Efficient Deep Learning Pipelines for Accurate Cost Estimations Over Large Scale Query Workload}


\author{Johan Kok Zhi Kang}
\email{johan.kok@u.nus.edu}
\orcid{??}
\affiliation{%
  \institution{National University of Singapore}
  \city{Singapore}
}

\author{Gaurav}
\email{gaurav@grab.com}
\orcid{??}
\affiliation{%
  \institution{GrabTaxi Holdings}
  \city{Singapore}
}

\author{Sien Yi Tan}
\email{sienyi.tan@grab.com}
\orcid{??}
\affiliation{%
  \institution{GrabTaxi Holdings}
  \city{Singapore}
}

\author{Feng Cheng}
\email{feng.cheng@grab.com}
\orcid{??}
\affiliation{%
  \institution{GrabTaxi Holdings}
  \city{Singapore}
}

\author{Shixuan Sun}
\email{sunsx@comp.nus.edu.sg}
\orcid{??}
\affiliation{%
  \institution{National University of Singapore}
  \city{Singapore}
}

\author{Bingsheng He}
\email{hebs@comp.nus.edu.sg}
\orcid{??}
\affiliation{%
  \institution{National University of Singapore}
  \city{Singapore}
}

\begin{abstract}
The use of deep learning models for forecasting the resource consumption patterns of SQL queries have recently been a popular area of study. With many companies using cloud platforms to power their data lakes for large scale analytic demands, these models form a critical part of the pipeline in managing cloud resource provisioning. While these models have demonstrated promising accuracy, training them over large scale industry workloads are expensive. Space inefficiencies of encoding techniques over large numbers of queries and excessive padding used to enforce shape consistency across diverse query plans implies 1) longer model training time and 2) the need for expensive, scaled up infrastructure to support batched training. In turn, we developed \textit{Prestroid}, a tree convolution based data science pipeline that accurately predicts resource consumption patterns of query traces, but at a much lower cost. 
We evaluated our pipeline over \HBS{19K Presto OLAP queries from Grab,} on a data lake of more than 20PB of data. Experimental results imply that our pipeline outperforms benchmarks on predictive accuracy, contributing to more precise resource prediction for large-scale workloads, yet also reduces per-batch memory footprint by 13.5x and per-epoch training time by 3.45x. We demonstrate direct cost savings of up to 13.2x for large batched model training over Microsoft Azure VMs.
\end{abstract}


\maketitle

\section{Introduction}
\label{introduction}
    \input{content/introduction}

\section{Background \& Related Work}
\label{background}
    \input{content/background}

\section{The curse of diversity \& scale} 
\label{case_study}
    \input{content/case_study}

\section{Design and Implementation}
\label{solution}
    \input{content/solution}

\section{Experimental Evaluation} \label{experiment_results}
    \input{content/experiment_results}



\section{Conclusion}
\label{conclusion}
    \input{content/conclusion}

\section{Acknowledgements}
\label{acknowledgements}
    \input{content/acknowledgements}


\bibliographystyle{plainnat}
\bibliography{citations/bibliography}

\label{appendix}
    \input{content/appendix}

\end{document}

%% file: content/introduction.tex
Present trends in big (OLAP) data query engine design have shown two key features; compatibility with cloud-based infrastructure and the adoption of a decoupled compute-storage paradigm. This shift introduces complexities that makes the use of analytical models to forecast a query's expected resource consumption increasingly difficult. In turn, there has been a rising adoption in using deep learning models \cite{marcus2019neo,kipf2018learned,ortiz2018learning} for such a task. These models, when trained over features extracted from prior executed queries, can yield good results in predicting the resource consumption of new, unseen queries. 

\begin{figure}[htb]
    \includegraphics[width=0.8\linewidth]{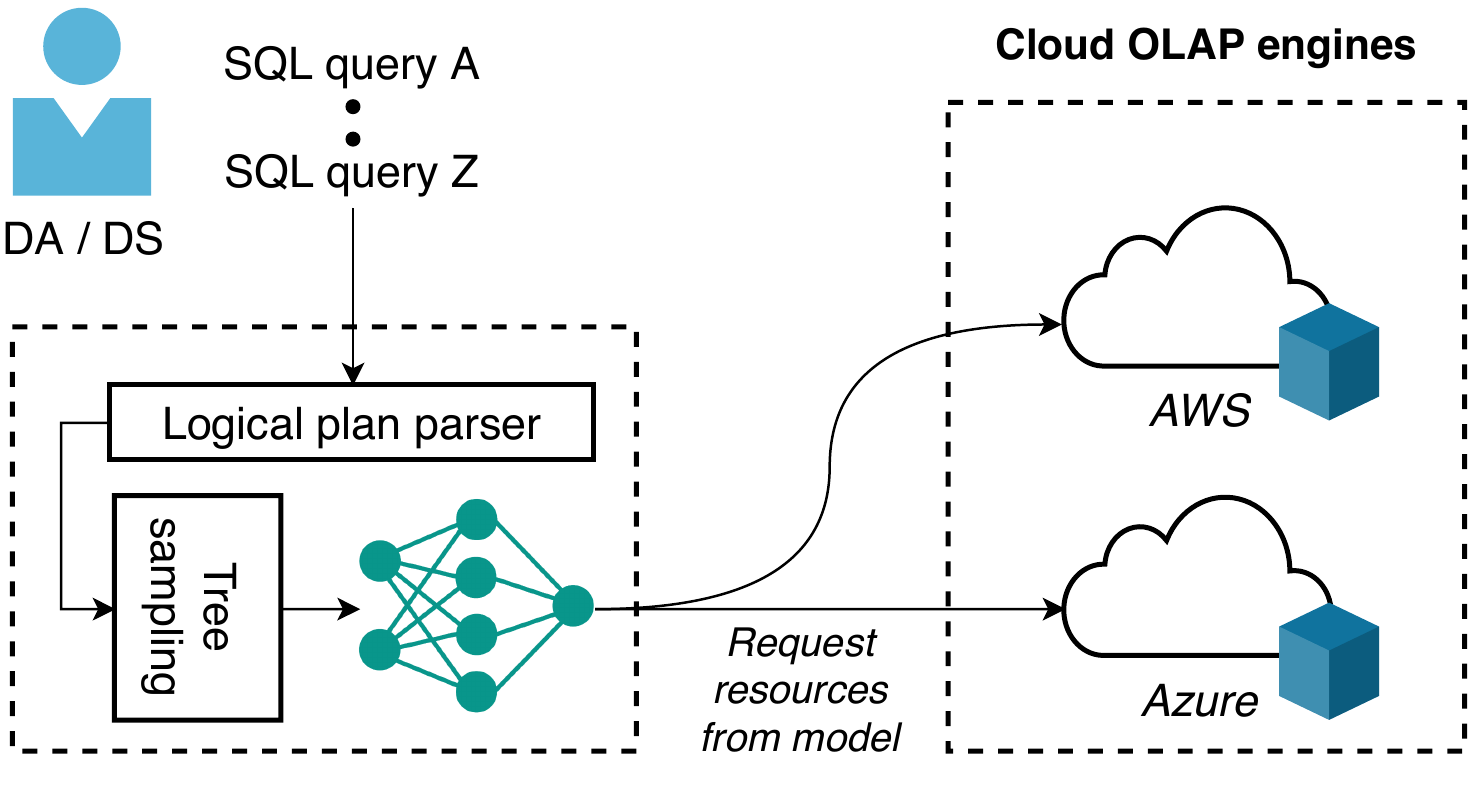}
    \caption{Integration of Prestroid for resource deployment}
    \label{fig:deployment_pipeline}
\end{figure}

Our work is heavily inspired by use cases from \HBS{Grab}, a large ride hailing company managing a data lake of more than 20PB in the cloud. As the business grows and diversifies, queries to our data lake are expanding both in volume and in variety. We record hundreds of thousands of queries issued over multiple Presto \cite{sethi2019presto} clusters powered by cloud infrastructure. Monthly cloud expenditures form a large portion of expenses, thus having a systematic and accurate framework for forecasting a query's resource utilization will entail huge cost savings \cite{gupta2008pqr,he2010comet}. Such a framework may be represented as an end-to-end pipeline in Fig \ref{fig:deployment_pipeline}. Incoming queries are first parsed for their features before being routed through a deep learning model. The model predicts the resources needed by the query and these resources are created for query execution. Such a system assures that resources allocated to run a query are neither excessive which incurs cost, nor insufficient at the risk of a query violating its service level agreement (SLA). 

To date, researchers have explored various deep learning pipelines for query-resource prediction. We review those works in Section \ref{background}. Herein, we observed two issues with a direct application of these pipelines. The first is that pipeline models are costly to be trained over the query patterns from \HBS{Grab}, in large variety and volume. Our experiments showed that a single model training over 19K queries can cost as much as \$76.25 for batch sizes of 256. The second is that model updates need to be done in perpetuity to keep up with the rapid evolution of the business. This amounts to a spiralling cost that could offset any savings gained from having a systematic resource allocation framework. To the best of our knowledge, there is limited work focused on making deep learning pipelines practical for companies managing data lakes at similar scale.

Our work is directed towards improving the cost efficiency for deep learning model training over a query's logical plan. We focus on optimizing the state-of-the-art tree convolution (Tree CNN) based pipelines. Present pipelines \cite{marcus2019neo, marcus2020bao} apply Tree CNN over full query plans and have poor per-batch memory footprint and long epoch runtime. In turn, we propose \textit{Prestroid}, a pipeline that addresses these problems and is cheaper for training. We denote our contributions below.

\begin{compactitem}[$\bullet$]
    \item We present a case study over SQL queries issued over \HBS{Grab's} data lake. Our study revealed the presence of a large number of distinct query predicates and a wide disparity in query plan sizes. We show why direct applications of present deep learning pipelines are 1) not space-efficient for encoding and 2) creates excessive padding that mandates the use of higher tier and more expensive GPUs for training.
    
    \item We present the components of Prestroid, consisting of a sub-tree convolution model, a Word2Vec model for controlling plan level embedding and a novel sub-sampling algorithm for decomposition of large query plans whilst preserving breadth level information for tree convolution. Prestroid is cost-efficient as it achieves better forecast accuracy at lower per-batch memory footprint and faster epoch runtime. 
    
    \item We demonstrated experimentally that our pipeline enables model training cost reduction of 2x and 13.2x at batch sizes of 32 and 256 respectively. Moreover, Prestroid achieves better predictive accuracy than state-of-the-art, which helps the resource provisioning for large-scale workloads in \HBS{Grab}.
    
    \item We publicly avail our \HBS{Grab-Traces} and TPC-DS dataset\footnote{\HBS{Plans to open source our dataset are on the way and we will add the dataset URL in due course of time.}}. To the best of our knowledge, \HBS{Grab-Traces} is the largest available industry-based dataset of query plans.
\end{compactitem}


%% file: content/background.tex

\subsection{Cloud based resource utilization}
Majority of cloud platforms offer pay-as-you-use resource type customization to meet the varied demands of customers \cite{vaquero2011dynamically}. These resources are packaged as a tiered set of on-demand  virtual machines (VMs) with different cores, memory \& storage capacities and pricing. A collection of VMs forms a \textit{cluster} and are the workhorses of big data query engines \cite{sethi2019presto, bisong2019google} on the cloud. As it is possible to add an indefinite number of VMs to a cluster (termed as \textit{scale out}), developing a good resource forecasting framework for projected query workload would enable the selection \cite{sen2020autotoken, jyothi2016morpheus} of just the right combination of VMs to meet demands at cost optimal pricing.

\subsection{State-of-the-art}
Analytical models \cite{sen2018characterizing,wu2013predicting,rajan2016perforator,huang2017top} attempt to shed light on the internal mechanisms of a query engine by modelling various aspects, such as data access and workload scheduling. Such models have been developed extensively for transactional based systems such as MySQL and Postgres or distributed processing engines such as Hive and Hadoop. However, such models are highly specific to a single engine and are hard to develop.

Machine learning models approach the problem differently. Earlier models explored simple techniques, such as KCCA \cite{ganapathi2009predicting}, for sub-space mapping of queries-resources, or standard regression analysis \cite{popescu2012same}. In recent years, there has been a shift towards deep learning models such as feed forward network \cite{kipf2018learned,ortiz2018learning}, Tree CNN \cite{marcus2019neo}, RNN based networks \cite{ortiz2019empirical,sun2019end} or reinforcement learning \cite{marcus2020bao,marcus2019towards}. These models were designed to capture the inherent complexities of OLAP queries to which simpler models failed to do so. 

\vspace{-2mm}

\subsection{Query feature extraction}
Deep learning models are trained and evaluated over numerical inputs. This implies the need to formulate methods that translate plain SQL into their vectorized representations. 

\textbf{SQL parsing} - A simple approach would be to parse a query string entirely to aggregate key features that represents the query. Gnapathi et al. \cite{ganapathi2009predicting} proposed 9 distinct features that characterizes an SQL query, where as Makiyama et al. \cite{makiyama2015text} suggested representing a query as a collection of the weighted frequencies of its individual word tokens.

\textbf{Logical plan parsing} - To venture further, aggregations may be done over a query's logical plan structure. A logical plan is directed acyclic graph (DAG) representation of the operations needed to be fulfilled before the final table can be materialized. Each operation is represented as a node in the DAG. Expressing a query as such allows deeper insights into the execution sequence taken by the query engine that may not be attained through plain text parsing. Such were the works proposed by \cite{akdere2012learning,akdere2011case,ganapathi2009predicting}, in which the authors represented a query as vector aggregation of specific operations within a plan. In modern database engines such as Presto or MySQL, obtaining a query's logical plan can be easily achieved using the "\textit{EXPLAIN <text>}" key word without the need to execute the query. 

\textbf{O-T-P encoding} - The \textit{Operator-Table-Predicate} encoding format has been adopted by many state-of-the-art work in the field of deep learning for query-resource forecasting \cite{marcus2019neo, kipf2018learned, ortiz2019empirical}. At its core, a distinction is made between the categories of \textit{Operators}, which are wildcards representing key operations such as joins or projections, \textit{TABLES} indicating the scanned tables and \textit{PREDICATES} indicating the conditions over which data is filtered. Different encoding techniques may be applied within each category and the resultant combination of \{O,T,P\} is the feature representation of that query. 

Such encoding techniques (with slight variants) were adopted in \cite{kipf2018learned,sun2019end,ortiz2019empirical,marcus2019neo} and worked well for a good variety of models, from simple feed forward networks \cite{kipf2018learned}, recurrent neural networks \cite{ortiz2019empirical,sun2019end} and Tree CNN networks \cite{marcus2019neo}. We adopt the O-T-P encoding approach in this work by first casting a query into its logical plan, before re-casting into a binary tree comprising only of O/T/P nodes. 

\subsection{Tree CNN based models}
SQL featurization, as a standalone, may fall short of adequately representing a query as they fail to account for the order of executions captured within the plan sequence \cite{mou2014convolutional}. In the context of database engines, the choice of plan may significantly impact run time performance \cite{armbrust2015spark,marcus2020bao,park2020quicksel}. Such sequence order sensitivity implies the need to develop models that are able to differentiate plan level spatial arrangements, in order to maximize predictive accuracy.

Tree CNNs are one such model. They draw inspiration from pre-existing convolution networks applied over images or graphs. The pioneering works of Mou et al. \cite{mou2014convolutional} have inspired the application of Tree CNN based models in the field of query plan selection \cite{marcus2019neo} and natural language classification \cite{bui2017cross,dq2019bilateral}. Tree CNN networks aggregate information between parent and children by sliding and pooling triangular kernels breadth first across each node, thereby capturing the positional ordering of operators within a plan. The reader is encouraged to review \cite{mou2014convolutional} for a deeper understanding.

%% file: content/case_study.tex
In this section, we present our analysis of sample query plans from \HBS{Grab} and highlight problems with excessive padding and dealing with a large, distinct set of query predicates. 

\subsection{An industry case study}
\label{industry_case_study_sec}
\begin{figure}[htb]
    \includegraphics[width=0.9\linewidth]{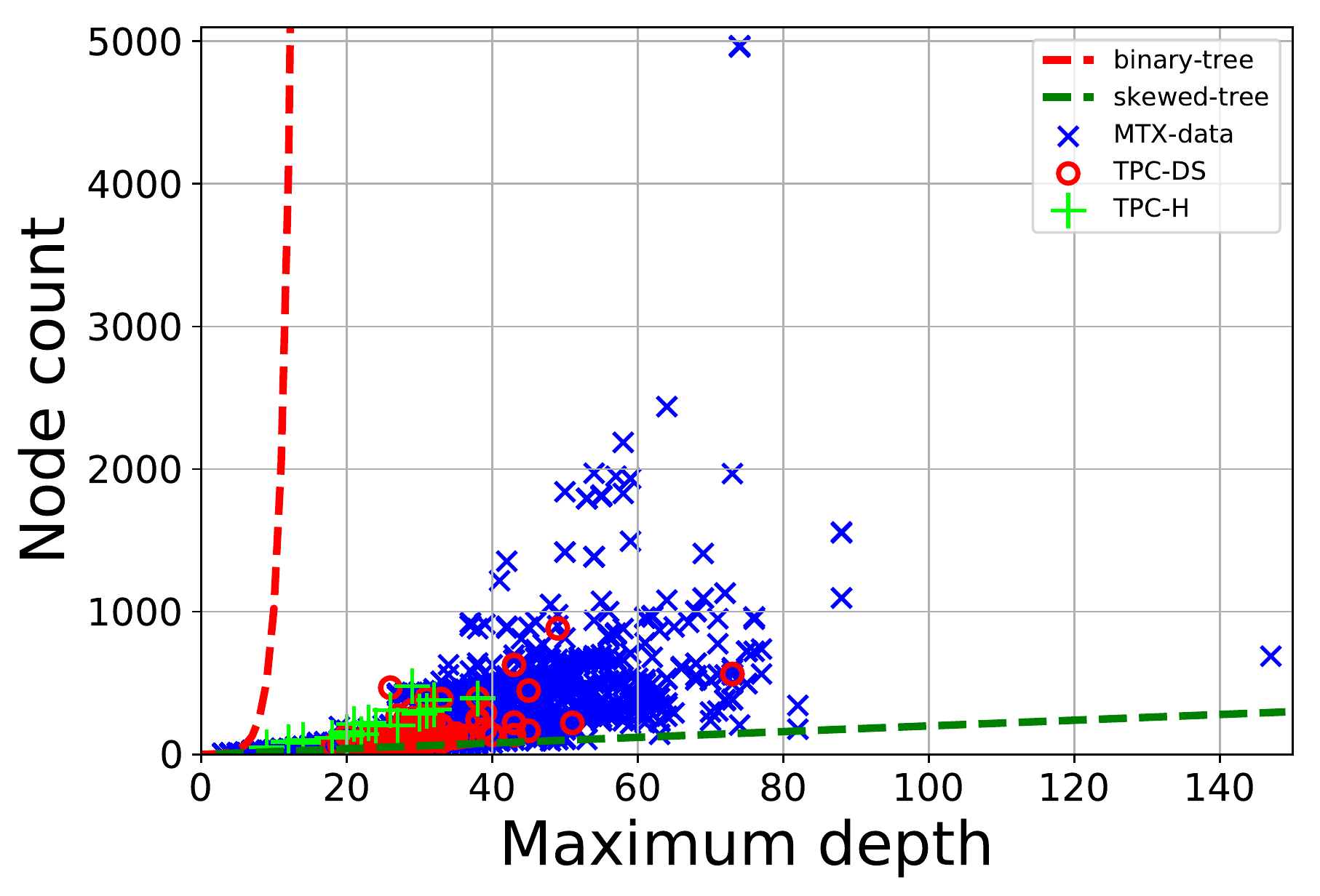}
    \caption{Contrast of 245,849 logical plan samples from \HBS{Grab} alongside 103 TPC-DS \& 22 TPC-H publicly available samples. We showed theoretical plots for skewed trees and balanced binary trees for reference.}
    \label{fig:logical_plan_distribution}
\end{figure}


\HBS{Grab is Southeast Asia's leading superapp that that provides everyday services such as mobility, deliveries (food, packages, groceries), mobile payments and financial services to millions of Southeast Asians. Rapid expansion across the region} has resulted in queries to our data lake that are vastly different in characteristic and resource needs. We plotted a sample of 245,849 logical plans, obtained over 2 months, on their node count and maximum depth in Fig \ref{fig:logical_plan_distribution}. Here, the maximum depth refers to the largest distance between the root and any leaf node. As reference, we contrasted them with theoretical plots for balanced binary and skewed trees (left-deep trees with only 1 child). An observation was that majority of the plans straddled in between both plots, indicative of plan diversity within the sample set.

\textbf{Distinction from public datasets} - We contrast the plans from \HBS{Grab} with publicly available TPC-DS \& TPC-H plans in Fig \ref{fig:logical_plan_distribution}. \textit{Notably, these plans covered a smaller range of distribution relative to the plans from \HBS{Grab}, implying that the latter was richer in quantity and the span of plan sizes}. To quantify, the maximum plan (size, depth) observed was (477, 38) for TPC-H, (883, 73) for TPC-DS and (4969, 321) for \HBS{Grab}. 

This distinction highlights an important area of research that is yet to be explored. To the best of our knowledge, most research fail to address the issues that surfaces in training Tree CNN based deep learning pipelines over a large and diverse set of logical plans. We attribute one possibility to the fact that publicly available query patterns lack both the quantity and permutations needed to replicate the scenario present in \HBS{Grab}. Addressing these issues is a critical step forward in bridging the gap between research and practical applications to the industry.  

\textbf{Dynamism of query patterns} - The performance of a deep learning model is highly dependent on the training dataset. As such, training frequency should be contingent on how rapidly new data is introduced. In order to quantify this rate, we sampled 373K Presto queries from \HBS{Grab} across 1 month span and extracted all tables required by the queries. We asked ourselves if the model was used to predict for a subsequent window of W days, what is the percentage of tables in the new queries that the model has not encountered.

\begin{table}[h]
    \begin{tabular}{||c c c c c c||}
        \hline
            W  & 1 & 3 & 5 & 7 & 9 \\
            \hline \hline
            \% & 1.65 & 4.76 & 7.64 & 9.27 & 12.18 \\
        \hline
    \end{tabular}
    \caption{Percentage of new tables that a model has not seen over the next W days window.}
    \label{tab:table-concept-drift}
    \vspace{-4mm}
\end{table}

\vspace{-2mm}
 We are observing a high rate of growth in tables within the company as seen in Table \ref{tab:table-concept-drift}. This motivates the frequent re-training of our models. For example, at W = 9, the model has been used to predict new queries over the next 9 days. In doing so, the model suffers from a high degree of inaccuracy, given that 12.18\% of tables scanned are new tables that the model has not been trained over. We therefore recommend the daily re-training of our models in practice.


\subsection{0-padding for dimensional consistency}
It is common practice to implement \textit{NULL}, or "0-padding" in order to reconcile irregularities in input data dimensions. For \HBS{Grab}, these irregularities appear due to having both large and small query plans in the training data. Theoretically, 0-padding will not impact model training performance as a null input does not affect weight updates. Yet 0-padding introduces redundant information in the model \cite{wu2017new} that has consequences. 

For a given training batch size, excessive 0-padding will lead to an increase in overall per-batch memory footprint. This leads to longer data transfer time between CPU-GPU for each epoch cycle and more computations needed over the data, resulting in longer per-epoch runtime. More importantly, scarce GPU memory bandwidth will be exhausted for models with multiple layers as the GPU has to retain all intermediate data to compute back propagation gradients. To \HBS{Grab}, the implications are two folds. Machine learning practitioners
either do not have the flexibility to tune models over a wide range of batch sizes, which may lead to sub-optimal model performances \cite{keskar2019large}, or they have to scale out their hardware to more expensive GPU tiers on the cloud, which is a cost concern.

One technique to avoid unnecessary 0-padding is down sampling. This problem has been largely explored in the field of image processing \cite{marin2019efficient,Vu_2018_ECCV_Workshops}. Unfortunately, such cannot be said for the field of deep learning over tree based structures. 

\subsection{Surge in query predicates}
A query predicate defines a condition to be applied over transformations within each stage of a query plan. In the case of conditional filters, predicates are represented as a set consisting of \{Columns, Comparison operator, Filter values\} \cite{sun2019end}.

Our analysis of the query patterns in \HBS{Grab} revealed that while the number of tables being queried are few, the number of unique predicates were very large relative to publicly available datasets\footnote{Based on 19,876 sampled queries from Grab in our training dataset, the number of unique predicate counts may extend as much as 30,707. In contrast, 5,153 TPC-DS queries, generated from 81 templates, yields a count of 1,450.}. This is understandable, given that a single table can have multiple columns for performing filters or joins and that predicates may vary in complexities according to business rules. In turn, we highlight the flaws of existing encoding techniques applied to these predicates. 


\textbf{1-Hot \cite{kipf2018learned}} - This may cause a sizeable increase in the encoded vector's length, creating sparse vectors of a single 1 and remaining 0s, which occupies a large chunk of encoding space. 

\textbf{Value normalization \cite{woltmann2019cardinality}} - 
Normalization is used to constrain filter values to a (0,1) range suitable for training. This technique works mostly for integer and floats. For strings, dictionary encoding may be used to cast it to an integer. Unfortunately, such technique do not work for predicate columns and must be coupled with others as discussed.

\textbf{R-vectors \cite{marcus2019neo}} - The R-vector representation was proposed by Marcus et al. in an attempt to capture the semantic relations between column values in a database. A Word2Vec model is trained, for encoding newly materialized tables for each query, by first treating each row in a table as a sentence and each column as a token. While R-vectors enable compact feature representations of queries, they are costly for deployment. Prior execution of each query is needed to materialize the table for encoding. This clearly does not work well for hundreds of thousands of queries, where each may take hours to complete. 

\subsection{Summary}
In summary, our observations of the scale and diversity of query workloads in \HBS{Grab} surfaces several problems with existing deep learning pipelines. Firstly, the need to reconcile both large and small query plans begets excessive 0-padding. Secondly, present encoding techniques for large query dataset are space inefficient. These factors amount to an increase in per-batch memory footprint and induces unnecessary computations. As consequence, model training has to be done over longer horizons and on scaled out GPU machines, which is costly overall. Directly addressing these problems will yield improvements to the cost efficiencies of such pipelines.


%% file: content/solution.tex
Here we present a deep dive into Prestroid's data pipeline design for addressing the challenges in Section \ref{case_study}. We present an overview of our end-to-end model training process before explaining how we reduced per-batch memory footprint through minimizing node level encoding and 0-padding. For simplicity, we focus on single objective learning in which the model has to predict how much \textit{total CPU time} a query consumes. 


\begin{figure*}[htp]
    \centering
    \begin{tabular}{@{}c@{}}
        \includegraphics[width=0.8\linewidth]{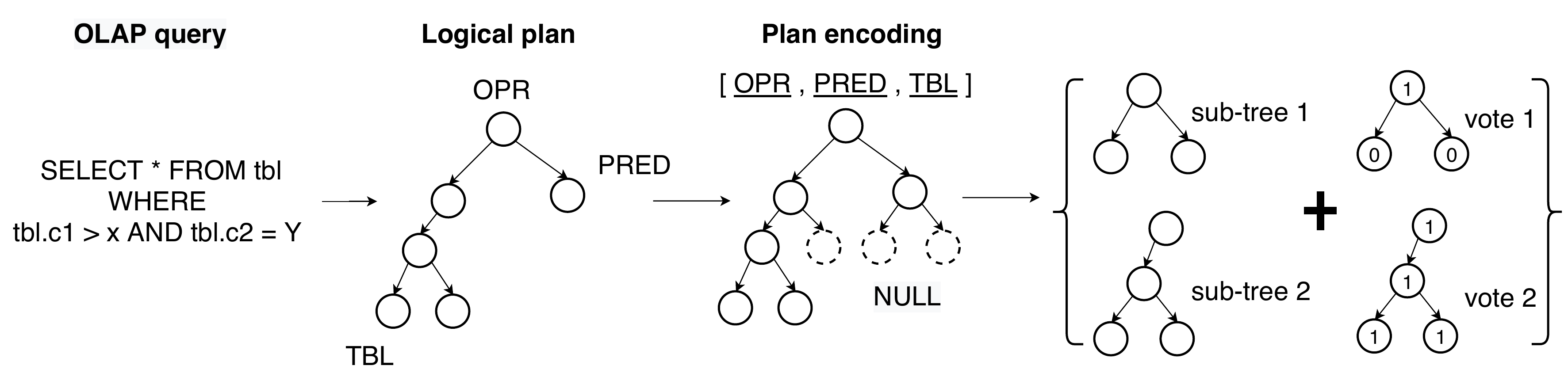}
        \\[\abovecaptionskip]
        \small (a) Query parsing and sub-tree decomposition.  
    \end{tabular}
    
    \begin{tabular}{@{}c@{}}
        \includegraphics[width=0.8\linewidth]{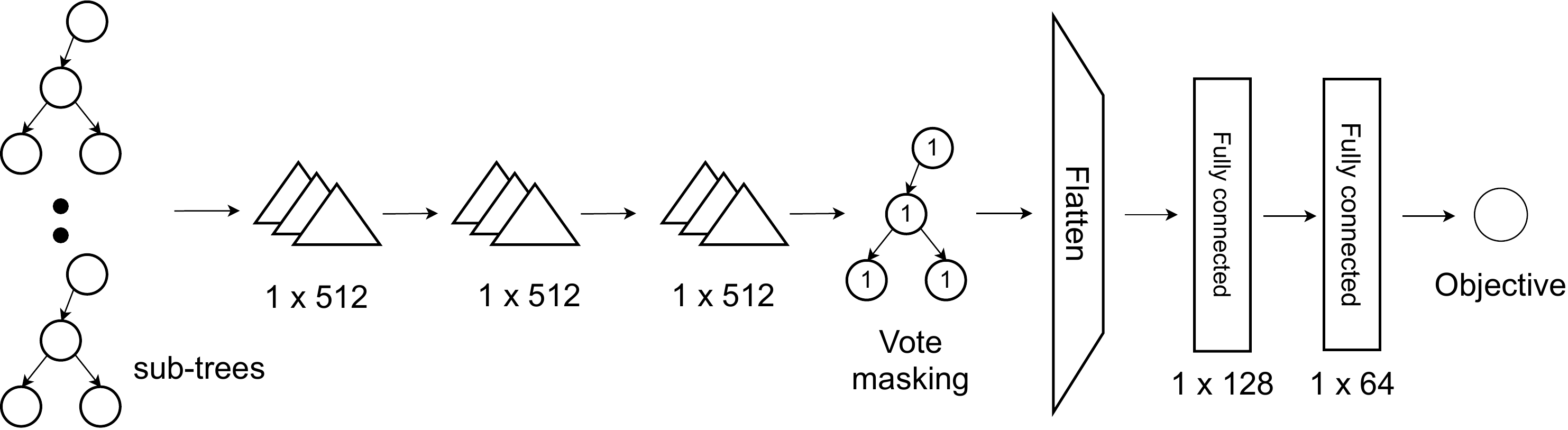}%
        \\[\abovecaptionskip]
        \small (b) Sub-tree model network architecture
    \end{tabular}
    
    \caption{Prestroid data pipeline used for model training \& prediction}
    \label{fig:data_pipeline}
    \vspace{-4mm}
\end{figure*}

\vspace{-1mm}

\subsection{Building the data pipeline}
\label{building_the_data_pipeline}
Prestroid consists of a data pre-processing phase and a model training phase, illustrated in Fig \ref{fig:data_pipeline}. Firstly, an incoming query is decomposed into its logical plan and further re-cast using the \textit{O-T-P} framework. We apply the following rules:

\begin{compactitem}[$\bullet$]
    \item For a non-join node $N_I$, set its type to be an OPR. Create a right child as type PRED with the predicate value. The left child is untouched.
    
    \item  For a join node $N_J$, set its type to be an OPR. The left \& right children are untouched.
    
    \item For a leaf node $N_L$ (table scans), set its type to be an OPR. Create a left child as type TBL with the table scanned as value. Create a right child as $\varnothing$.
    
    \item Transform the resultant tree into a binary tree by adding $\varnothing$ to any node with fewer than 2 children.
\end{compactitem}

\textbf{Plan encoding} - OPR \& TBL nodes are collected separately and 1-Hot encoded whilst PRED nodes are encoded using our model in Section \ref{predicate_embedding}. We traverse each node in a tree and apply respective encoding in the [OPR, PRED, TBL] format. 


\textbf{Sub-tree model} - We apply our sub-sampling algorithm and select the first K sub-trees as representative features for a query. Our model uses 3 layers of CNN. We then apply bit masking and perform one-way dynamic pooling \cite{mou2014convolutional} over each sub-tree. Finally, we flatten all sub-trees into a single vector,  pass them through 2 dense layers with ReLU activation before a single layer with sigmoid activation as prediction.


\subsection{Learned predicate embedding} \label{predicate_embedding}

\textbf{Word2Vec model} - The goal here is to identify an n-dimensional feature space that enables control over each predicate embedding in a meaningful way. For example, the words "LONGITUDE" and "LATITUDE" appear frequently with each other in the queries that we sampled. We would expect them to be spatially closer in our feature space as compared to "DATAMART", which is used in a totally separate context. 

\begin{figure}[htb]
    \includegraphics[width=1\linewidth]{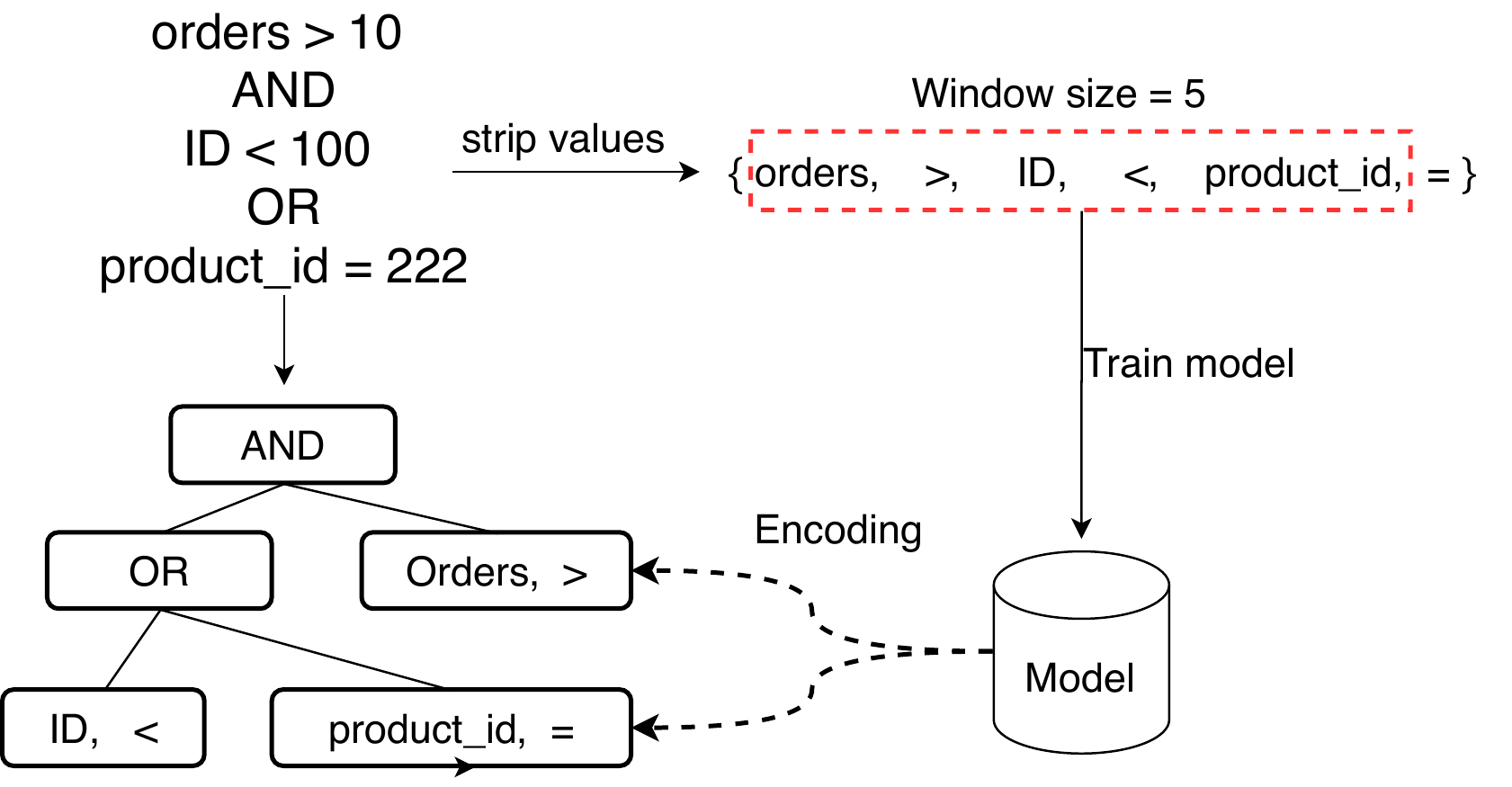}
    \caption{Illustration of our Word2Vec model training and encoding over a toy predicate example}
    \label{fig:predicate_flow}
\end{figure}

Such a problem has been explored extensively in the arena of natural language. Popular models proposed are Word2Vec models \cite{mikolov2013efficient} such as skip-gram or CBOW. We show that such models can be used for learning predicate representations based on logical plan extracts from queries. The key idea is to train our Word2Vec models over all predicate tokens with values omitted. To illustrate, consider the example in Fig \ref{fig:predicate_flow}. To train our model, all conjunctions and values from each predicate are first stripped off, leaving behind only the columns and comparison operators. We then train our token sets using the Word2Vec model offered by Python's Gensim package. We ran the model using a window size of 5, minimum token count of 10 and a range of feature sizes. Tuning the feature size allows us to control the encoding space for predicates.

\textbf{Handling conjunctions} - After obtaining our trained encoding, we cast our predicate into a tree where the nodes are either conjunctions (AND or OR) or a single predicate clause. For the latter, we encode each word token and take the overall average as the node level encoding. We then apply MIN feature pooling over all children nodes for AND conjunctions and MAX feature pooling for OR conjunctions, following prior works \cite{sun2019end}. 

\textbf{Out of vocabulary tokens} - One approach to address out of vocabulary tokens encountered during pipeline deployment is to follow a hierarchy of updates. For example, in order to encode unseen predicates, we first search for PRED nodes within the query and take their average features. If not, we take the average of all tokens in the query. If all else fails, we take the average encoding for all PRED nodes in the global set. This simple approach works well in our experiments.

\subsection{Sub-tree sampling}
The goal here is to decompose a large query tree into smaller sub-trees. Thereafter, by careful selection of the top K representative sub-trees, we are able to reduce the overall 0-padding needed to our model. Algorithm 1 denotes the pseudo code for our sub-tree sampling algorithm. In contrast to naive breadth first or depth first pruning, our sub-sampling algorithm ensures that information needed during Tree CNN is preserved.

The crux of the algorithm lies in the observation that, given a root node \textbf{R}, node count limit \textbf{N} and \textbf{C} layers of convolution, R only requires information present in its children up till C levels below. Hence, the rough workings of our algorithm are as such.

\begin{compactitem}[$\bullet$]
    \item Starting from R (depth 0), we consider the possibility growing our sub-tree at incremental depths of 1. At each depth, all children must be materialized. 
    \item Where the sub-tree has node counts exceeding N at depth D, we regress back to depth D-1 and prune the tree breadth first. 
    
    \item All nodes up till depth (D-C-1) have complete information and are allowed to vote\footnote{A vote is simply a bit masked value that is applied after all convolution layers the model. A vote of 1 is assigned only to nodes that have complete information and can be used as valid signals during post-convolution.}. We set their votes to 1. Nodes beyond depth (D-C-1) have their votes set to 0. 

    \item The algorithm is repeated for all nodes at depth (D-C).
\end{compactitem}

\begin{algorithm}
\SetAlgoLined
\SetKwInOut{Input}{input}
\Input{
    \ N node limit, C convolution layers, R root node
}
\vspace{1mm}

\SetKwInOut{Output}{output}
\Output{
    \ $[S_n]$ sub-sample binary trees, $[V_n]$ - Votes
}
\vspace{1mm}

\SetKwInOut{Constraints}{constraints}
\Constraints{
    \ N $>$ $2^{C+1}$ -1 
}
\vspace{2mm}

def \textit{getNodes}(R: root node, D: depth): \\
\ \ \ \ returns all nodes until depth D, including R

\vspace{1mm}

$\varnothing \leftarrow$ FIFO queue $Q_f$, $[\ ] \leftarrow$ S, V \\
$Q_f$.enqueue(R) \\

\While{not $Q_f$.empty}{
    node = $Q_f$.pop() \\
    candidates = $[\ ]$ \\
    depth = 0 \\
    \vspace{3mm}
    
    \# Terminate when we hit leaf nodes or limit \\
    \While{len(candidates) $\leq$ N} {
        prior\_candidates = candidates \\
        depth += 1 \\
        candidates = getNodes(node, depth) \\
        \uIf{len(candidates) == len(prior\_candidates)} {
            \# No new children nodes to discover \\
            break
        }
    }
    \vspace{3mm}
    
    sub\_tree = prior\_candidates \\
    sub\_tree\_count = len(sub\_tree)
    
    \uIf{len(candidates) == len(prior\_candidates)} {
        \# All nodes are valid for complete tree \\
        votes = [1] * sub\_tree\_count
    }
    \uElse {
        eligible = len(getNodes(node, depth -C -1)) 
        \\
        vote\_nodes = [1] * eligible + [0] * (sub\_tree\_count - eligible) 
        \\
        $Q_f$.enqueue(getNodes(node, depth -C) 
        \\
    }
    
    S.append(sub\_tree) \\
    V.append(votes) \\
}
return (S, V)
\caption{Sub-tree sampling algorithm}
\end{algorithm}

Our algorithm reduces the use of 0-padding by enabling the user to tune the values of N / K, which reduces the input Tensor's dimension to the model. Current implementation treats each plan as a binary tree and enforces the rule N > 2$^{C+1}$ -1.

%% file: content/experiment_results.tex
\begin{table*}[h]
    \begin{subtable}[h]{0.45\textwidth}
        \centering
        \begin{tabular}{||c c c||}
            \hline
            Models & Epoch & MSE \\
            \hline \hline
            Log bins & - & 96.91 \\
            SVR & - & 106.16 \\ 
            M-MSCN & 78 & 66.35 \\ 
            WCNN-100 & 55 & 50.35 \\
            WCNN-250 & 55 & 50.90 \\
            Full-100 & 52 & 50.82 \\
            Full-300 & 51 & 48.16 \\
            Prestroid (15-9-300) & 49 & 49.23 \\
            Prestroid (32-11-200) & 41 & \textbf{46.09} \\
            \hline
        \end{tabular}
        \caption{Performances on \HBS{Grab-Traces} dataset}
        \label{tab:grab_benchmarks}
     \end{subtable}
     \hfill
     \begin{subtable}[h]{0.45\textwidth}
        \centering
        \begin{tabular}{||c c c||}
            \hline
            Models & Epochs & MSE \\
            \hline \hline
            Log bins & - & 58.09 \\
            SVR & - & 58.97 \\ 
            M-MSCN & 17 & 145.91 \\ 
            WCNN-100 & 15 & 100.62 \\
            WCNN-250 & 29 & 103.05 \\
            Full-50 & 75 & 58.33 \\
            Full-100 & 69 & 55.60 \\
            Prestroid (15-47-50) & 46 & \textbf{46.61} \\
            Prestroid (32-32-100) & 49 & 47.24 \\
            \hline
       \end{tabular}
       \caption{Performances on TPC-DS dataset}
       \label{tab:tpc_ds_benchmarks}
     \end{subtable}
     \caption{Recorded MSE errors ($minutes^2$) for best performing Prestroid sub-tree (N-K-$P_f$) models, Prestroid full tree models and respective comparisons. We also included the highest observed epoch at convergence out of all 3 runs.}
     \vspace{-8mm}
     \label{tab:mse_results}
\end{table*}

Here we designed 3 experiments to evaluate Prestroid relative to selected benchmarks. \textit{Exp 1} - We assess how well Prestroid performed in terms of MSE score. All models were trained for 3 rounds, with average MSE scores taken from the best performing iterations. In each round, early stopping was employed to prevent model over fitting.  \textit{Exp 2} - We evaluate the accuracy of Prestroid in forecasting suitable resources quantities. \textit{Exp 3} - We evaluate per-batch memory foot print, required epoch training time and training costs for Prestroid sub-tree models over cloud based infrastructures. 

\subsection{Experimental Setup}
\textbf{Infrastructure} - Our models were trained using Tensorflow \cite{abadi2016tensorflow} over Azure. We used the NC\_V3 series powered by NVIDIA Tesla V100 GPUs with 16GB memory. Exp 1 \& 2 was conducted on NC12s\_V3 cluster with 2 GPUs. Exp 3 was conducted over NC6s\_V3 / NC12s\_V3 / NC24s\_V3 clusters fully utilizing all 1 / 2 / 4 GPUs available. 

\textbf{Dataset} - Our dataset was based on 2 different sources.

\textit{TPC-DS}: We generated a total of 5,153 unique queries with 81 unique templates from the TPC-DS Hive dataset. We filtered all queries with total CPU time between 1 - 60 min. All queries were executed on Presto at a scale factor of 10. We applied log transformation followed by min-max normalization over all recorded total CPU time to constrain all training values in between 0 - 1. We used a split ratio of 8 / 1 / 1 for training / validation / testing. Splitting was done at the template level.

\textit{\HBS{Grab-Traces}}: We curated a 2 month sample of query traces from \HBS{Grab}. These queries were executed across multiple Presto clusters in deployment. Only successfully executed queries were selected. We first filtered all queries with total CPU time between 1 - 60 min. We then applied log transformation followed by min-max normalization to constrain all training values in between 0 - 1. The resultant dataset contained 19,876 queries split into 8 / 1 / 1 ratio for training / validation / testing.

\textbf{Comparisons} - All deep learning model comparisons were trained using ADAM \cite{kingma2014adam} optimizer, batch size of 64 and the Huber loss function, unless stated otherwise.

\begin{compactitem}[$\bullet$]
    \item \textit{Log binning} \cite{dougherty1995supervised} - We split all query plans by their node counts into B log bins. The average total CPU timing is taken within each bin and used for inference. We used this as a naive benchmark for comparison with other models. Experimentally, we found that the optimal values for B were 1000 \& 20 for \HBS{Grab-Traces} \& TPC-DS dataset respectively.
    
    \item \textit{SVR} \cite{ganapathi2009predicting} - A support vector regression (SVR) model is trained using direct query parsing and plan operator instance counts. We omitted plan operator cardinalities as part of the feature vector. We found that the best performing models used a polynomial kernel of degree 4 and sigmoid kernel of degree 3 for \HBS{Grab-Traces} \& TPC-DS dataset respectively.
    
    \item \textit{Modified MSCN} \cite{kipf2018learned} - We modified the multi-set convolutional network (M-MSCN) for our task. Although MSCN was built for cardinality estimation, its design principles were based on Deep Sets \cite{zaheer2017deep}, which we argue allows the network to be generalized for query-cost regression estimation. We used 0-padding for input consistency and set dropout to 5\%. We set learning rate of 1$e^{-3}$ and 256 perceptron units per layer for \HBS{Grab-Traces} and 1$e^{-4}$ and 24 for TPC-DS.
    
    \item \textit{WCNN} \cite{zolaktaf2020facilitating} - We followed the Word Convolution (WCNN) implementation and optimal hyper-parameters as reported. We explored the use of \{100, 250\} kernels for each of the \{3, 4, 5\} sliding window convolution filters. We used a token embedding layer of dimensions 100 in our network. Dropout was set at 50\%, batch size at 16 and learning rate at 1$e^{-3}$ for \HBS{Grab-Traces} and 1$e^{-4}$ for TPC-DS.
    
    \item \textit{Prestroid (Full-$P_f$)} - To show the gains from using sub-trees, we implemented Prestroid over full query plans without any tree pruning. This model is similar to the tree convolution segment of \cite{marcus2019neo}. Here, $P_f$ represents the feature size chosen from our Word2Vec model. We explored the range of $P_f$ $\in$ \{100, 200, 300\} and \{50, 100\} for \HBS{Grab-Traces} \& TPC-DS dataset respectively. We set learning rate of 1$e^{-4}$.
\end{compactitem}

\subsection{Hyper-parameter tuning} 
Prestroid exposes 3 new parameters that may be tuned for performance. They are 1) \boldmath{$P_f$} $\rightarrow$ predicate features size, 2) \textbf{K} $\rightarrow$ number of sub-trees chosen to represent a query and 3) \textbf{N} $\rightarrow$ max node count per sub-tree. We explored 2 variations of N $\in$ \{15, 32\}. For \HBS{Grab-Traces}, we explored $P_f$ $\in$ \{100, 200, 300\}, K $\in$ \{5, 9, 21\} where N = 15 and K $\in$ \{5, 11, 20\} where N = 32. For TPC-DS, we explored $P_f$ $\in$ \{50, 100\}, K $\in$ \{31, 43, 47\} where N = 15 and K $\in$ \{20, 28, 32\} where N = 32.

We used 512 / 512 / 512 CNN kernels and 128 / 64 perceptron units for \HBS{Grab-Traces}. For TPC-DS, we scaled the architecture down to 128 / 128 / 128 and 32 / 8. We set dropout as 10\% for kernel and bias weights, with batch normalization in between each dense layer. We used ADAM with learning rate of 1$e^{-4}$ and optimized for Huber loss. Herein, We refer to any future variations of Prestroid as Prestroid (N-K-$P_f$).

\textbf{Performance on \HBS{Grab-Traces}} - Our results on \HBS{Grab-Traces} dataset imply that sub-tree models have greater learning capacities than other state-of-the-art deep learning models. The optimal sub-tree configuration observed was (32-11-200). 

In comparison with full tree models, our sub-tree models have K times more features as inputs to the dense layer after convolution, unlike the former which collapses all plan level features into a single vector via dynamic pooling. Scaling up the inputs by K times enabled our sub-tree models to learn a richer set of mappings between query plans and cost estimates.

Surprisingly, WCNN showed comparable performances to Prestroid Full-100, implying that convolution models that operate directly on SQL strings are able to extract just as much information as compared full query plans. However, SQL strings do not reflect the true cost of how a query is executed. Take the case of a simple command "SELECT * FROM A, B, C". Whilst WCNN understands that multiple joins are performed for tables A, B \& C, the join ordering and the type of join used is hidden from the model. These details are decided by the query optimizer at runtime and can only be accessed at the plan level. Consequently, this limits the learning ability of WCNN as compared to Prestroid sub-trees, which we showed had better learning than full trees.

Finally, both Prestroid full and sub-trees generally fared better than SVR and M-MSCN. The latter models were trained over an aggregation of features from both query and logical plans, which caused valuable plan level information to be lost. In contrast, Prestroid was able to leverage these signals through the use of triangular kernels tuned to detect the spatial patterns between parent and children nodes.

\textbf{Performance on TPC-DS} - While we observed that Prestroid (15-47-50) yielded the lowest MSE score over the TPC-DS dataset, the shortage of query variations yields two interesting observations. The first is that simpler models (Log Binning \& SVR) showed comparable performance relative to deep learning models, a trend absent when benchmarked over \HBS{Grab-Traces}. We assert that the latter models are harder to train and require a broad range of query characteristics and training data size to outperform our naive baselines. Unfortunately, such features were absent in the TPC-DS dataset, which scarcely contains only 103 publicly available query templates.

The second is a sharp decline in the performance for WCNN. WCNN models are relatively heavy; WCNN-100 contains 363,301 trainable parameters whereas Full-100 contains 195,469 trainable parameters. This implies that it is easier for WCNN to overfit when limited training data is present.

\subsection{Resource allocation}
\begin{figure}[htb]
    \includegraphics[width=1\linewidth]{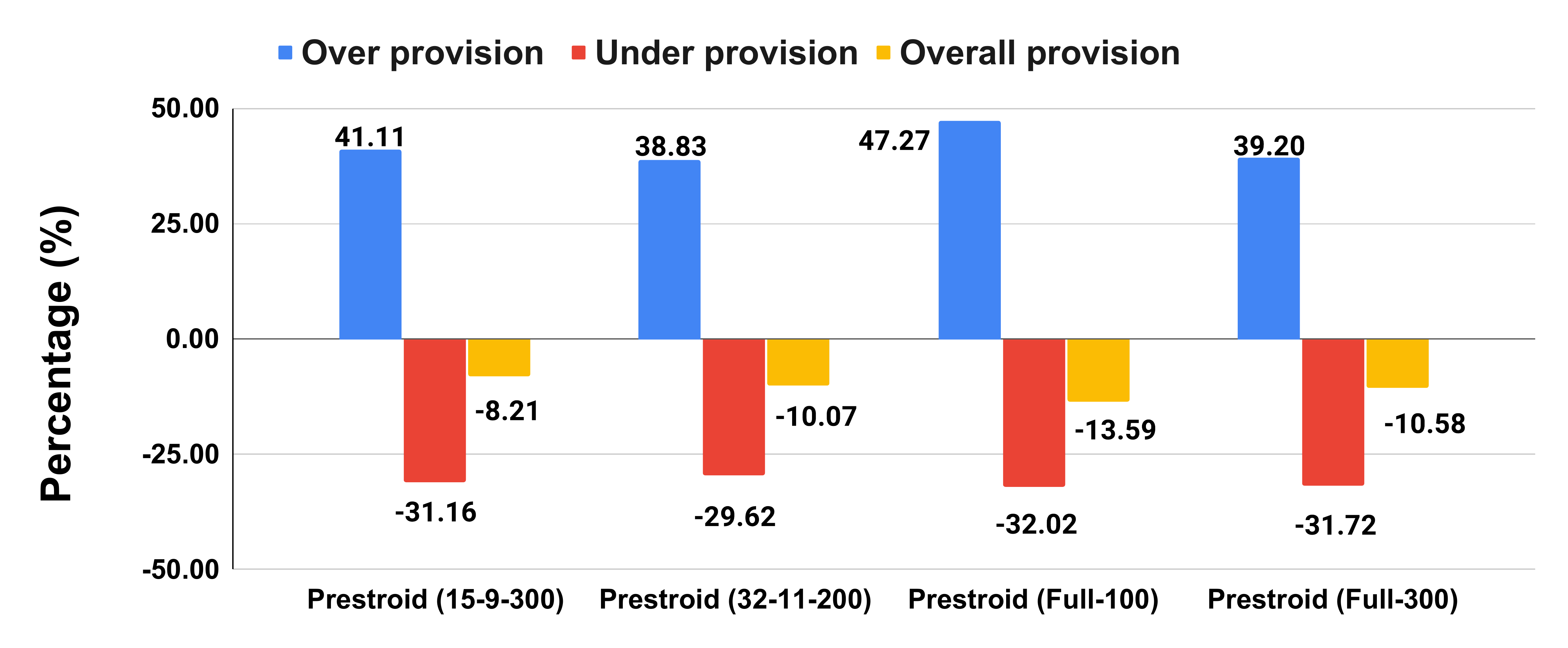}
    \vspace{-2mm}
    \caption{Percentage of cluster resources that were over / under allocated for \HBS{Grab} sampled query workloads. The lower the magnitude, the better the accuracy.}
    \label{fig:resource_allocation}
\end{figure}

\vspace{-4mm}

We designed an experiment to evaluate the resource allocation accuracy of Prestroid. Our results in Fig \ref{fig:resource_allocation} were based off a test data set of 1,987 query traces from \HBS{Grab-Traces}. We categorized our results into 2 groups: \textit{Over provisioned} \& \textit{Under provisioned}, with the intent of understanding by how much percentage of actual cluster resources did our sub-tree models over/under allocate to execute these queries. For example, queries which our model assigned excess CPU time were classified as Over provisioned. We compared Prestroid (15-9-300) \& (32-11-200), which were the best overall performers, against Prestroid (Full-100) \& (Full-300). Our observations were that Prestroid sub-trees tend to perform better for both over allocated and under allocated queries. Overall, all models generally under provisioned resources (see yellow bar), with Prestroid sub-trees achieving better resource allocation accuracy. This highlights the importance of our pipeline in enabling \HBS{Grab} to achieve optimal cost strategies for resource allocation.

\begin{figure}[htp]
    \centering
    \begin{tabular}{@{}c@{}}
        \includegraphics[width=0.9\linewidth]{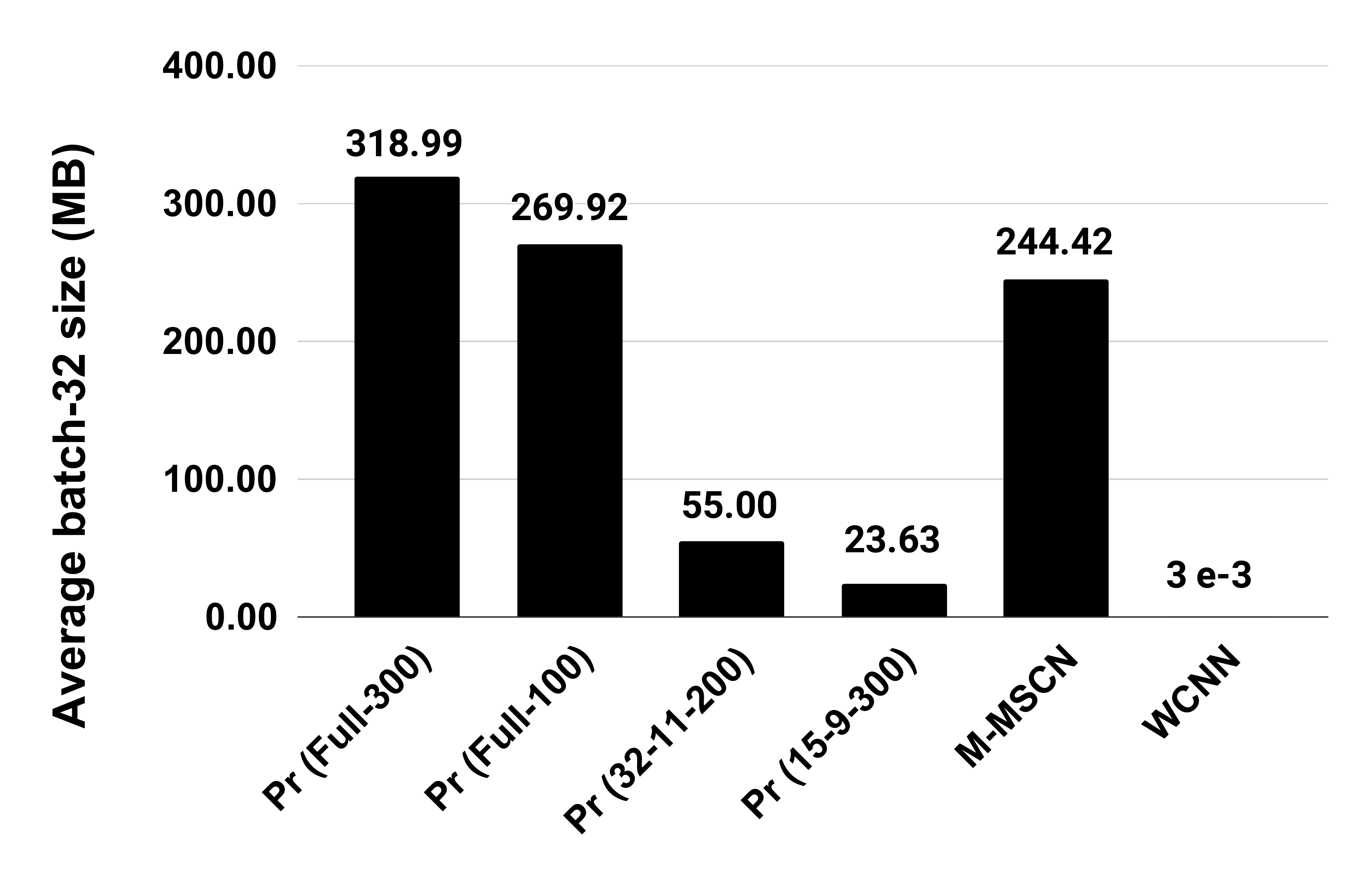}
    \end{tabular}
    \hfill
    \begin{tabular}{@{}c@{}}
        \includegraphics[width=0.9\linewidth]{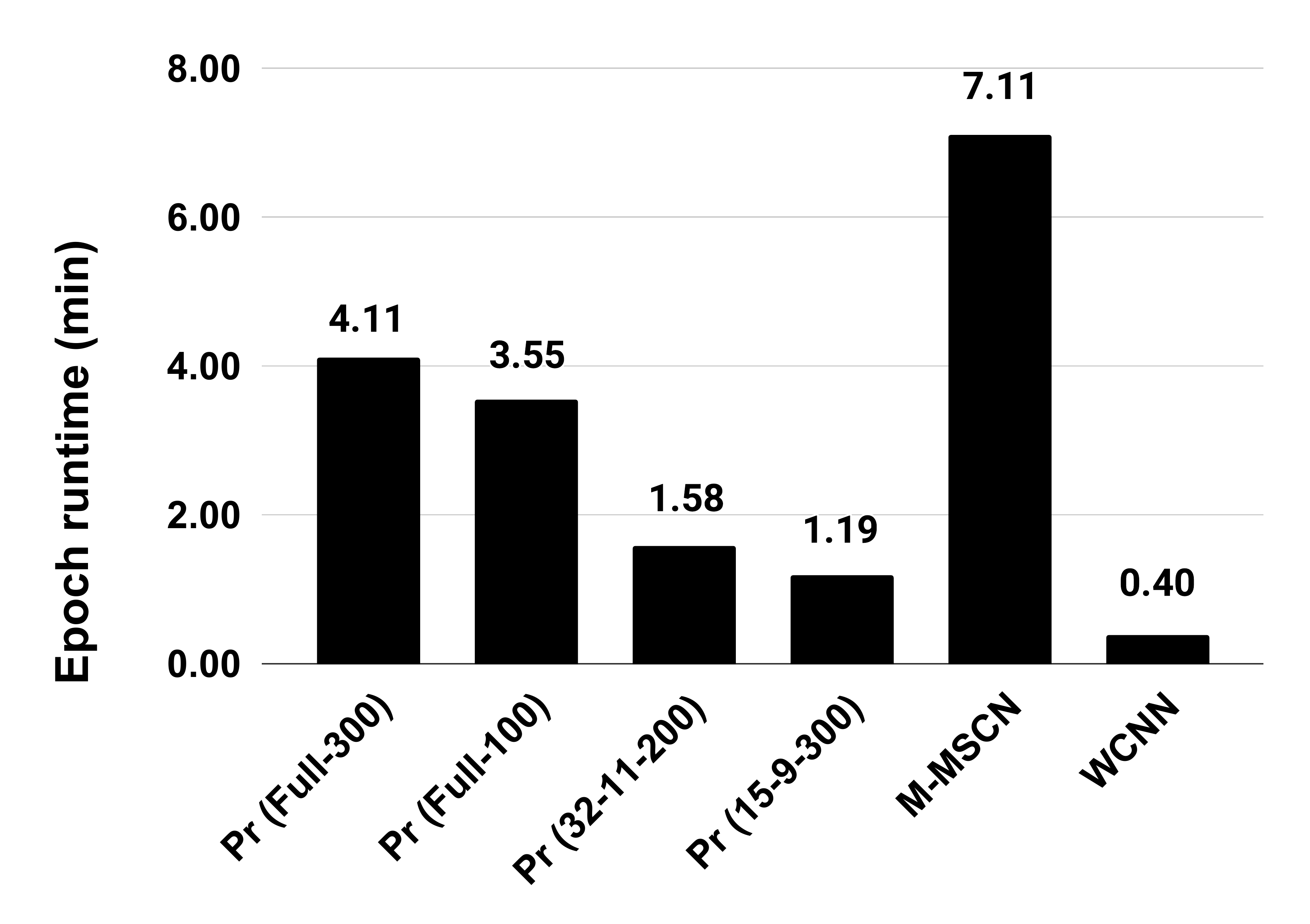}
    \end{tabular}
    \vspace{-6mm}
    \caption{Top: Average per-batch memory footprint (MB) for various models at batch size of 32 over \HBS{Grab-Traces} dataset. \textit{Pr} to denote Prestroid models. Bottom: Average epoch time (minutes) for various models.}
    \label{fig:hardware_consumption}
    \vspace{-8mm}
\end{figure}

\subsection{Optimizing hardware usage}
We studied how our encoding and sub-sampling approach improves batch efficiency and training speeds relative to full-tree models in Fig \ref{fig:hardware_consumption}. In turn, these improvements may be translated to cost savings over cloud resources. 

\textbf{Reduction in batch size} - The impact of 0-padding was significant given input size irregularities in \HBS{Grab-Traces} query plans. For a given batch size of 32, comparisons between Prestroid (15-9-300) \& (32-11-200) relative to Prestroid (Full-300) yields a reduction of 13.5x and 5.8x respectively. 0-padding was used to enforce dimensional consistency for inputs to all models. Padding for full tree models was based on the size of the largest tree present (1945 nodes). This meant padding a skewed distribution of small trees to the size of the largest. 

\textbf{Reduction in training time} - Reductions in input data size consequently leads to faster training times owing to improved bandwidth utilization for data transfer between CPU to GPU and fewer computations needed over the data. Comparisons between Prestroid (15-9-300) \& (32-11-200) relative to Prestroid (Full-300) yields speed up of 3.45x and 2.6x respectively. However, we also observed a disproportionate growth in epoch runtime for larger selections of K. We attribute this to an inefficiency in our current code, which uses Tensorflow's \textit{tf\_map} operator to perform sequential convolutions over each sub-tree. This limitation may be addressed in future work. 

\textbf{Other comparisons} - For completeness, we have included the per-epoch run time and batch sizes for M-MSCN and WCNN. For M-MSCN, a large number of distinct predicates coupled with variations in table, join and predicate sets per query produced sparse and large input tensors to the model, creating long epoch run times and high memory footprint. On the contrary, the use of a trainable token embedding layer in WCNN allowed us to minimize inputs to a single 1-D vector. This is highly efficient for speed and memory footprint reduction. Yet WCNN has shown sub-optimal performances relative to Prestroid sub-tree models in Table \ref{tab:mse_results}, which we argue is of first importance when designing query-cost estimation models.

\textbf{Cost Savings} - Finally, we evaluated the cost of model training using Azure's NC6s\_V3 / NC12s\_V3 / NC24s\_V3 clusters. We chose the lowest possible cost among all clusters that permitted training with a specified batch size. Each cluster contained 1 / 2 / 4 GPUs respectively and were priced at an hourly rate\footnote{Rates were up-to-date as of the time of writing and may change in future} of \$4.23 / 8.47 / 18.63. In multi-GPU clusters, data parallelism \cite{krizhevsky2014one} was employed to distribute batch workloads. We compared Prestroid (15-9-300) \& (32-11-200), which were the best overall performers, against Prestroid (Full-100) \& (Full-300). We assumed model training until the epochs denoted in Table \ref{tab:grab_benchmarks}.



\begin{figure}[htb]
    \includegraphics[width=1\linewidth]{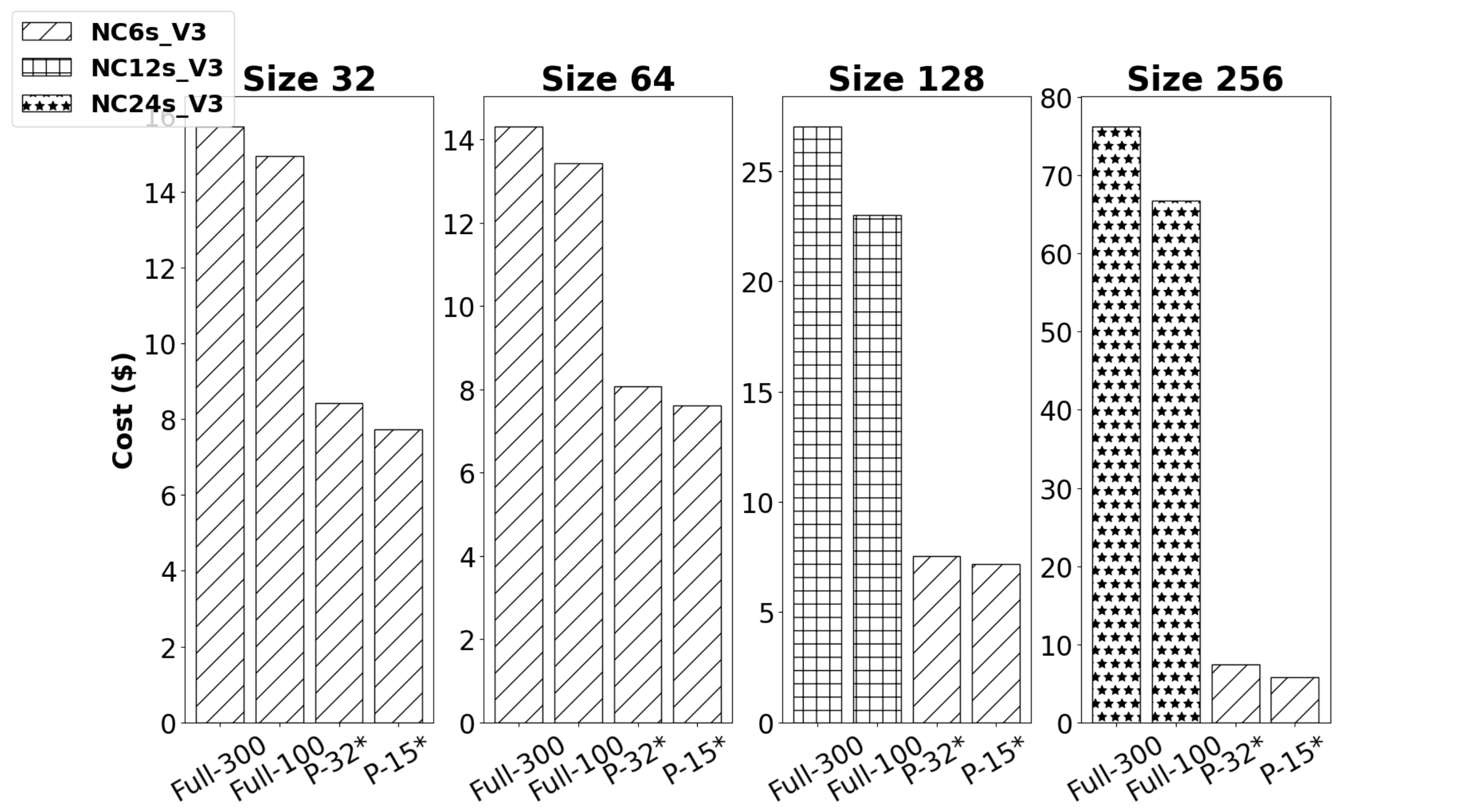}
    \vspace{-2mm}
    \caption{Lower bound of costs using Azure NC\_V3 clusters for training with varying batch sizes. P-15* \& P-32* denotes Prestroid  (15-9-300) \& (32-11-200) respectively}
    \vspace{-4mm}
    \label{fig:azure_cost}
\end{figure}

\textbf{Batch size costs} - \HBS{We observed diminishing cost returns due to communication overheads and a non-linear increase in pricing for multi-GPU clusters. Consequently, it is economically cheaper for training to be done over a single GPU.} Fig \ref{fig:azure_cost} suggests that our sub-tree models were cheaper to train across a range of batch sizes as they were faster and had lower memory footprint. For large batches, Full tree models had to be trained on multi-GPUs due to out of memory errors, while sub-tree models could still be trained on a single GPU. To quantify the impact of our sub-sampling approach, we observed training cost decline of \$76.25 to \textbf{\$5.79} in switching from Prestroid (Full-300) to Prestroid (15-9-300) for a batch size of 256.

%% file: content/conclusion.tex
This paper tackles the challenge of extending deep learning Tree CNN models for large and diverse queries in a cost efficient manner. Our sub-tree model exposes three levers which users can tune to control model accuracy, batch size and epoch training time. Careful selection of these parameters allows one to accelerate the training process over tens of thousands of query plans at lower memory footprint and high accuracy. This enables model training to be done faster and over cheaper cloud resources, leading to substantial cost savings. Although our experiments were conducted using data specific to \HBS{Grab}, we believe that the general techniques can be distilled and applied to any other company managing data lakes at a similar scale. 

%% file: content/acknowledgements.tex
\HBS{This work was funded by the Grab-NUS AI Lab, a joint collaboration between GrabTaxi Holdings Pte. Ltd. and National University of Singapore. We thank See-Kiong Ng, Hannes Kruppa and Rahul Penti for their support and advise.}

%% file: content/appendix.tex
\appendix

\section{Further discussions on MTX-Traces dataset}
\subsection{The long tail distribution in MTX-Traces}
\begin{figure}[htb]
    \includegraphics[width=\linewidth]{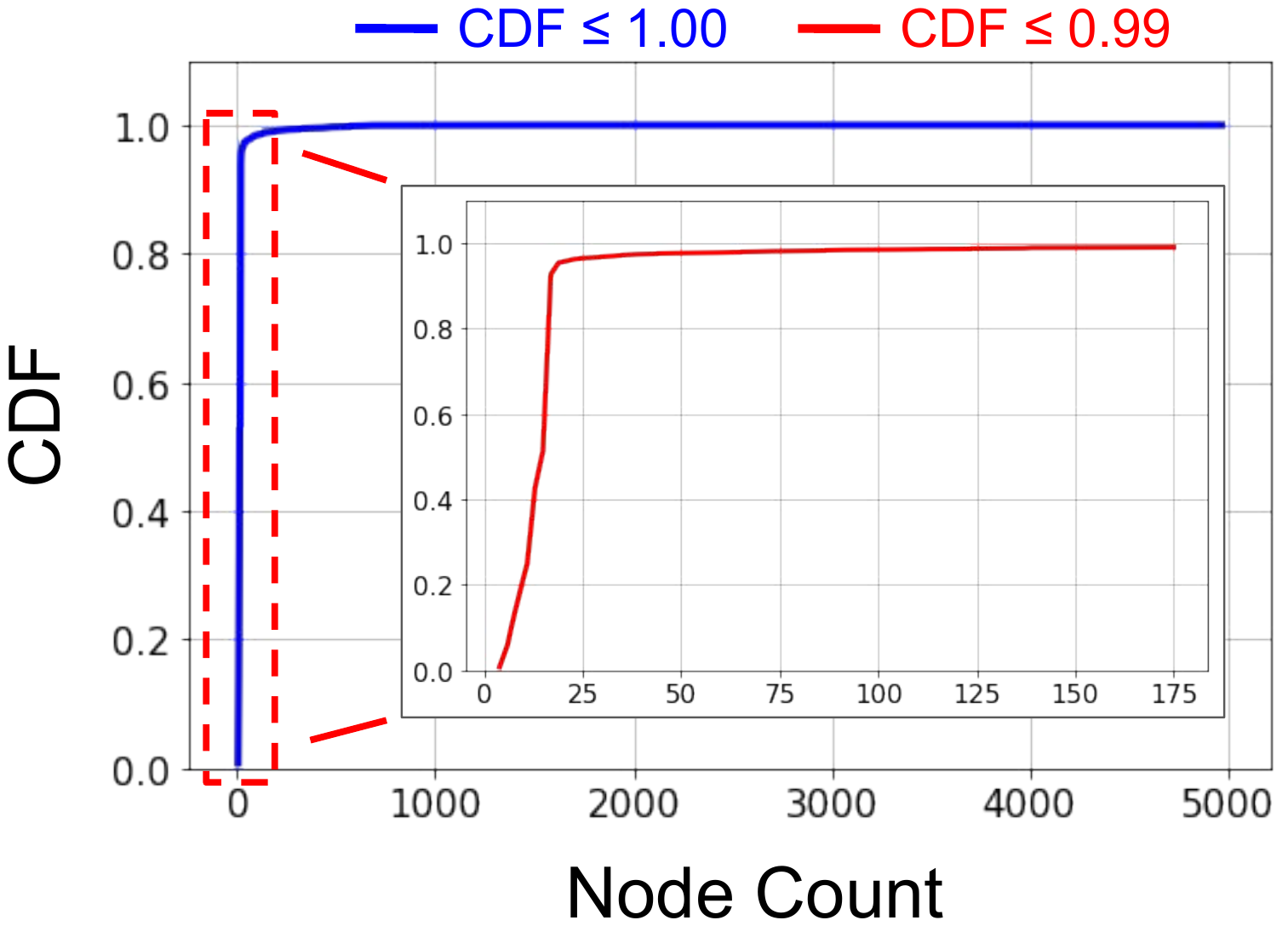}
    \caption{The long tail distribution of node count is evident over 245,849 logical plan samples from MTX.}
    \label{fig:cdf_node_counts}
\end{figure}

A standard practice in machine learning would be to remove the long tail distribution of data before model training, as these points are anomalous and may not represent the population at large. Indeed, analysis of our sample query plans revealed a highly skewed distribution of plan sizes, evident from Fig \ref{fig:cdf_node_counts}, that seemed to indicate the presence of strong outliers within the top 1 percentile. Yet we highlight the importance of these long tailed plans based on analysis of their overall resource consumption values. We selected a few cluster level resource types for our profiling. They were \textit{peak memory} recorded from query executions, \textit{total CPU time} across all cluster VMs and the \textit{input data size} ingested by each query. All these metrics were readily available from the Presto profiler. We observed that resource consumption for the top 1\% of queries was 23.7\%, 33.1\% and 40.2\% of cluster resources respectively. This meant that although the top 1 percentile of plans were small in numbers, they consumed a disproportionately large amount of cluster resources. It is important to expose our models to these plans in order to improve overall resource allocation framework.

\section{More experimental results}
\subsection{Discussion on scale out penalties}
\begin{figure}[htp]
    \centering
    \begin{tabular}{@{}c@{}}
        \includegraphics[width=1\linewidth]{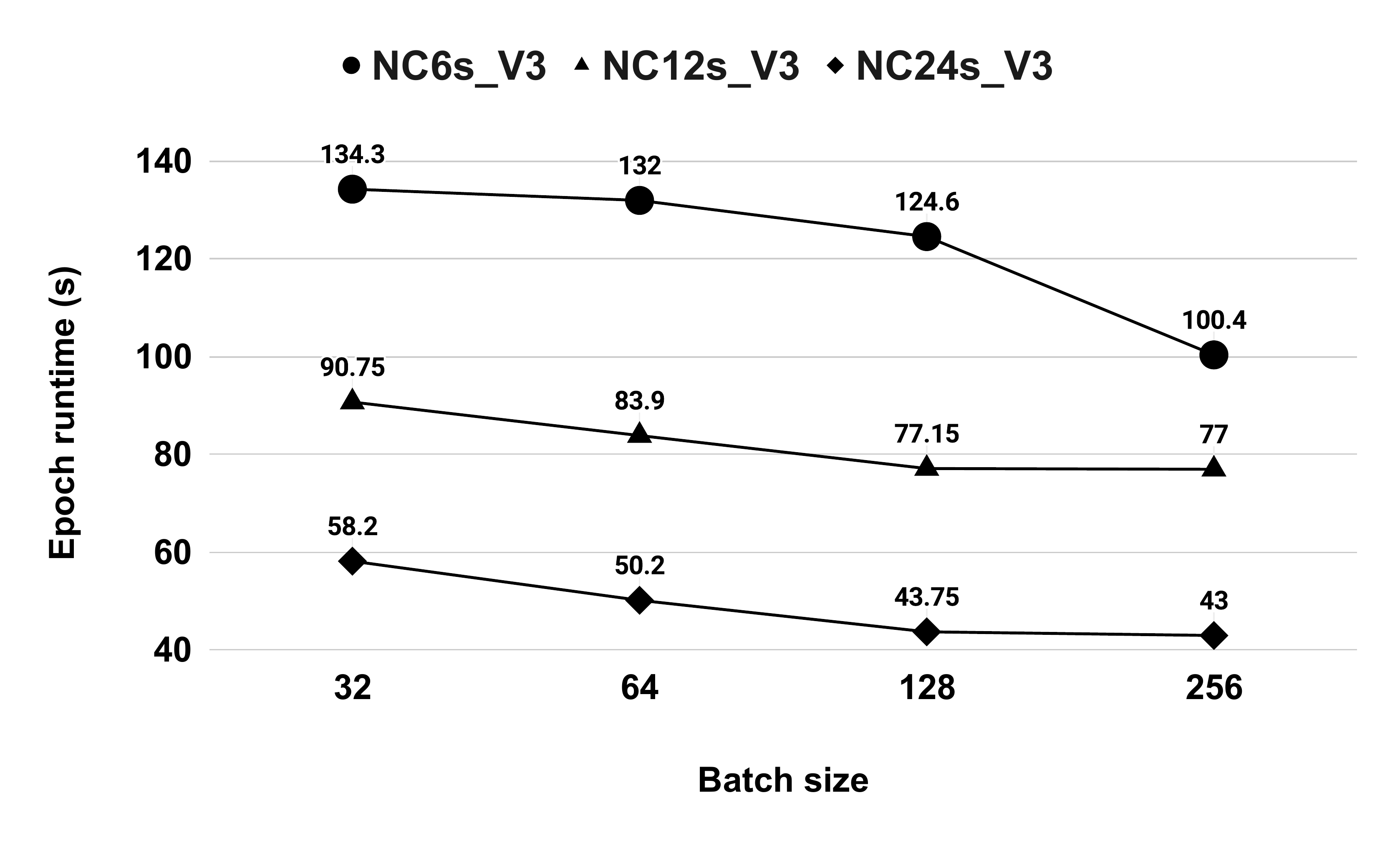}
    \end{tabular}
    \caption{Profiling  of  batch  size  against  epoch  runtime (s) for Prestroid (15-9-300)}
    \label{fig:scale_out_penalties_pr_15}
\end{figure}

\HBS{The default parallelism strategy employed by Tensorflow is data parallelism \cite{shallue2019measuring}. Conceptually, distributed model training is achieved by first replicating all model weights across each participating machine in the cluster. At each epoch, data is evenly sharded and distributed over all machines. Each machine computes its local weight updates via back propagation and sends the results to a common parameter server. The server aggregates all the weight updates and redistributes the new weights to each machine before further training.} 

\HBS{Unfortunately, the repercussions of such a strategy, when scaled across multiple machines, are two folds. Firstly, a large the number of machines participating in the training process yields a large synchronization cost, given that a single parameter server will be bandwidth bottle-necked by the communication of multiple weight updates asynchronously. Secondly, deep learning models with larger number of parameters tend to incur higher communication overheads, due to the need to transmit more model parameter values across the network in each epoch. Such is the case for Prestroid sub-tree models, which are relatively heavy compared to full trees and WCNN.} 

\HBS{In order to quantify the penalties of scaling out over our GPU clusters, we conducted an experiment in which we varied the batch sizes for Prestroid (15-9-300) whilst recording the average epoch runtime over Azure NC6s\_V3 / NC12s\_V3 / NC24s\_V3 clusters, with 1 / 2 / 4 GPUs respectively. We observed diminishing returns in scaling out training from single GPU to multi-GPU instances. Fig \ref{fig:scale_out_penalties_pr_15} shows a clear illustration of the scale out penalties incurred. For example, when trained using a batch size of 128, we observed speed ups of 1.62x / 2.85x vs the theoretical speed up of 2x / 4x when scaling from 1 GPU to 2 \& 4 GPUs respectively. Such penalties imply that cost savings attained via faster model training over N clusters of machines are unlikely to offset the (minimally) N fold increase in cost price, thereby making model training over scaled out cloud based resources less economical.}

\subsection{Inference timings}
\HBS{In order to evaluate the responsiveness of our models in a production setting, we tested inference timings based on a test set of 1,987 data over MTX-Traces dataset. We restricted our inference to only 1 NVIDIA Tesla v100 GPU and varied the inference batch size within the range of \{32, 64, 128, 256, 512, 1024\} for each model, such that the inference timing was optimal. We report the results in Table \ref{tab:inference_timings}}.

\begin{table}[h]
    \centering
    \begin{tabular}{||c c c||}
        \hline
        Models & Batch size & Timing (s) \\
        \hline \hline
        M-MSCN & 128 & 19.92 \\ 
        WCNN-100 & 512 & 4.91 \\
        WCNN-250 & 512 & 5.92 \\
        Full-100 & 64 & 15.44 \\
        Full-300 & 64 & 16.83 \\
        Prestroid (15-9-300) & 512 & 15.18 \\
        Prestroid (32-11-200) & 512 & 17.83 \\
        \hline
    \end{tabular}
    \caption{\HBS{Inference timings and optimal inference batch size used over MTX-Traces dataset}}
    \label{tab:inference_timings}
\end{table}

\HBS{We note that the inference timing for Prestroid sub-tree \& full-tree models are higher relative to WCNN models. This was due to the high computational requirements of Prestroid model architectures during the forward pass. Our sub-tree \& full-tree models have 512 filters at each layer of convolution, as compared to WCNN with 100 / 250 filters per layer. Convolution layers are compute intensive, as supported by the research by Krizhevsky et al. \cite{krizhevsky2014one}}. 

\HBS{In addition, since subsampling of query plans has enabled the reduction of input data size for our sub-tree models, we made attempts to scale up the batch size used for sub-tree model inference, to ensure maximum utilization of GPU resources. As a result, we were able to scale up the batch size for sub-tree models to 512 whereas Full-tree models were capped below a batch size of 128. However, one major source of inference bottleneck for Prestroid sub-trees models lies in the sequential computation of convolution kernels over each sub-tree using Tensorflow tf\_map operator. For large choices of K, as in the case of Prestroid (32-11-200), we observed long inference timings relative to Full tree models. This issue can be addressed in future improvements to our sub-tree model design}. 

\subsection{Error distribution}
\HBS{In order to evaluate the stability of model training, we repeated the training process for all models 3 times with early stopping. For each repetition, the best performing epoch was taken and MSE score computed. We provide training error standard distributions over MTX-Traces \& TPC-DS dataset in Table \ref{tab:std_error}}.

\begin{table}[h]
    \begin{subtable}[h]{0.45\textwidth}
        \centering
        \begin{tabular}{||c c||}
            \hline
            Models & Std \\
            \hline \hline
            M-MSCN & 0.41 \\ 
            WCNN-100 & 1.89 \\
            WCNN-250 & 1.27 \\
            Full-100 & 3.91 \\
            Full-300 & 0.78 \\
            Prestroid (15-9-300) & 1.34 \\
            Prestroid (32-11-200) & 1.92 \\
            \hline
        \end{tabular}
        \caption{Std error over MTX-Traces}
        \label{tab:grab_benchmarks}
     \end{subtable}
     \hfill
     \begin{subtable}[h]{0.45\textwidth}
        \centering
        \begin{tabular}{||c c||}
            \hline
            Models & Std \\
            \hline \hline
            M-MSCN & 16.23 \\ 
            WCNN-100 & 3.23 \\
            WCNN-250 & 0.48 \\
            Full-50 & 4.82 \\
            Full-100 & 3.09 \\
            Prestroid (15-47-50) & 6.89 \\
            Prestroid (32-32-100) & 10.75 \\
            \hline
       \end{tabular}
       \caption{Std error over TPC-DS}
       \label{tab:tpc_ds_benchmarks}
     \end{subtable}
     \caption{\HBS{Standard deviation (Std) errors ($minutes^2$) observed for all model training process over MTX-Traces \& TPC-DS dataset.}}
     \label{tab:std_error}
\end{table}

\HBS{We observed that the standard deviation of model training scores was, in general, higher for TPC-DS relative to MTX-Traces. We attribute this to the lack of query template variations and limited size of the TPC-DS dataset. Our TPC-DS dataset consists of 5K data points which were constructed from 81 out of 103 publicly available templates, with only the predicate fields varying between queries. Since all of our deep learning models stand to benefit from variations in query structure and training data size, it was unsurprising that we observed higher instability in model training results on the TPC-DS relative to MTX-Traces. Such observations promote the effectiveness of our MTX-Traces dataset as an industry realistic benchmark for the future development of deep learning models for query-cost estimation}. 

\subsection{Performance on time shifted dataset}
\HBS{Finally, we briefly compared the performances of our sub-tree \& full-tree models on a new dataset consisting of 780 Presto query data points, sampled from a 1 week period outside of the data range from our MTX-Traces query dataset. This exercise was done to understand Prestroid's performance when deployed over the scenario of a dynamically evolving datalake, as present in the case of MTX. We report the results in Table \ref{tab:time-shifted-performance}}.

\begin{table}[h]
    \begin{tabular}{||c c||}
        \hline
        Model & MSE \\
        \hline \hline
        Full-100 & 120.16 \\
        Full-300 & 123.67 \\
        Prestroid (15-9-300) & 125.39 \\
        Prestroid (32-11-200) & 129.62 \\
        \hline
    \end{tabular}
    \caption{\HBS{MSE scores ($minutes^2$) for best performing Prestroid sub-tree \& full-tree models after inference over a new 1-week sample of MTX dataset.}}
    \label{tab:time-shifted-performance}
\end{table}

\HBS{In such situations, we observed a significant model performance degradation, given the presence of new query plans which our model has never encountered before. We partially justify this hypothesis with our results from Table \ref{tab:table-concept-drift}, to which we observed significant deviations of known tables that our models have been trained over as our window size W grows. In this case, the introduction of new tables not only contributes to unseen TBL tokens but also PRED tokens used to represent the new table columns, which leads to further model inaccuracies. As such, we will explore large training data sets and more frequent re-training, given that this work has significantly reduced per-model training cost over cloud based resources}.